\documentclass{article}

\usepackage{microtype}
\usepackage{graphicx}
\usepackage{subfigure}
\usepackage{booktabs} 
\usepackage{nicefrac}       
\usepackage{microtype}      
\usepackage{float}
\usepackage{hyperref}
\usepackage{algorithm}

\usepackage{amsmath}
\usepackage{graphicx}
\usepackage{xcolor}
\usepackage{gensymb}
\usepackage{stmaryrd}
\usepackage{amssymb}
\usepackage{bbm}

\newcommand{\program}{\rho}

\newcommand{\library}{\mathcal L}
\newcommand{\libparams}{\theta_\library}
\newcommand{\eval}[1]{\llbracket{#1}\rrbracket}
\newcommand{\joint}{J}
\newcommand{\jointparams}{\theta_\joint}
\newcommand{\optimaljointparams}{\theta_{\joint^*}}

\newcommand{\desc}{d_t}
\newcommand{\translation}{\mathcal T}

\DeclareMathOperator*{\argmax}{arg\,max} 
\newcommand{\probability}{\text{P}} 

\usepackage[accepted]{icml2021}

\icmltitlerunning{Leveraging Language to Learn Program Abstractions and Search Heuristics}

\begin{document}

\twocolumn[
\icmltitle{Leveraging Language to Learn Program Abstractions and Search Heuristics}

\begin{icmlauthorlist}
\icmlauthor{Catherine Wong}{mit}
\icmlauthor{Kevin Ellis}{cornell}
\icmlauthor{Joshua B. Tenenbaum}{mit,cbmm}
\icmlauthor{Jacob Andreas}{mit}
\end{icmlauthorlist}

\icmlaffiliation{mit}{MIT}
\icmlaffiliation{cbmm}{Center for Brains, Minds and Machines (CBMM) - MIT}
\icmlaffiliation{cornell}{Cornell University}

\icmlcorrespondingauthor{Catherine Wong}{\hbox{catwong@mit.edu}}

\icmlkeywords{Program Synthesis, Program Induction, Natural Language, Grammars, Machine Learning, ICML}

\vskip 0.3in
]



\printAffiliationsAndNotice{
} 

\begin{abstract}
Inductive program synthesis, or inferring programs from examples of desired behavior, offers a general paradigm for building interpretable, robust, and  generalizable  machine learning systems. 
Effective program synthesis depends on two key 
ingredients: a strong library of 
functions from which to build programs, and an efficient search strategy for finding programs that solve a given task.
We introduce LAPS (Language for Abstraction and Program Search), a technique for
using {\em natural language annotations} to guide joint learning of libraries and neurally-guided search models for synthesis. 
When integrated into a state-of-the-art library learning system (DreamCoder), 
LAPS produces higher-quality libraries and
improves search efficiency and generalization on three domains  
-- string editing, image composition, and abstract reasoning about scenes --   
even 
when no natural language hints are available at test time.
\end{abstract}

\section{Introduction}\label{section-introduction}
\begin{figure*}[t]\label{banner}\vskip 0pt  \includegraphics[width=\textwidth]{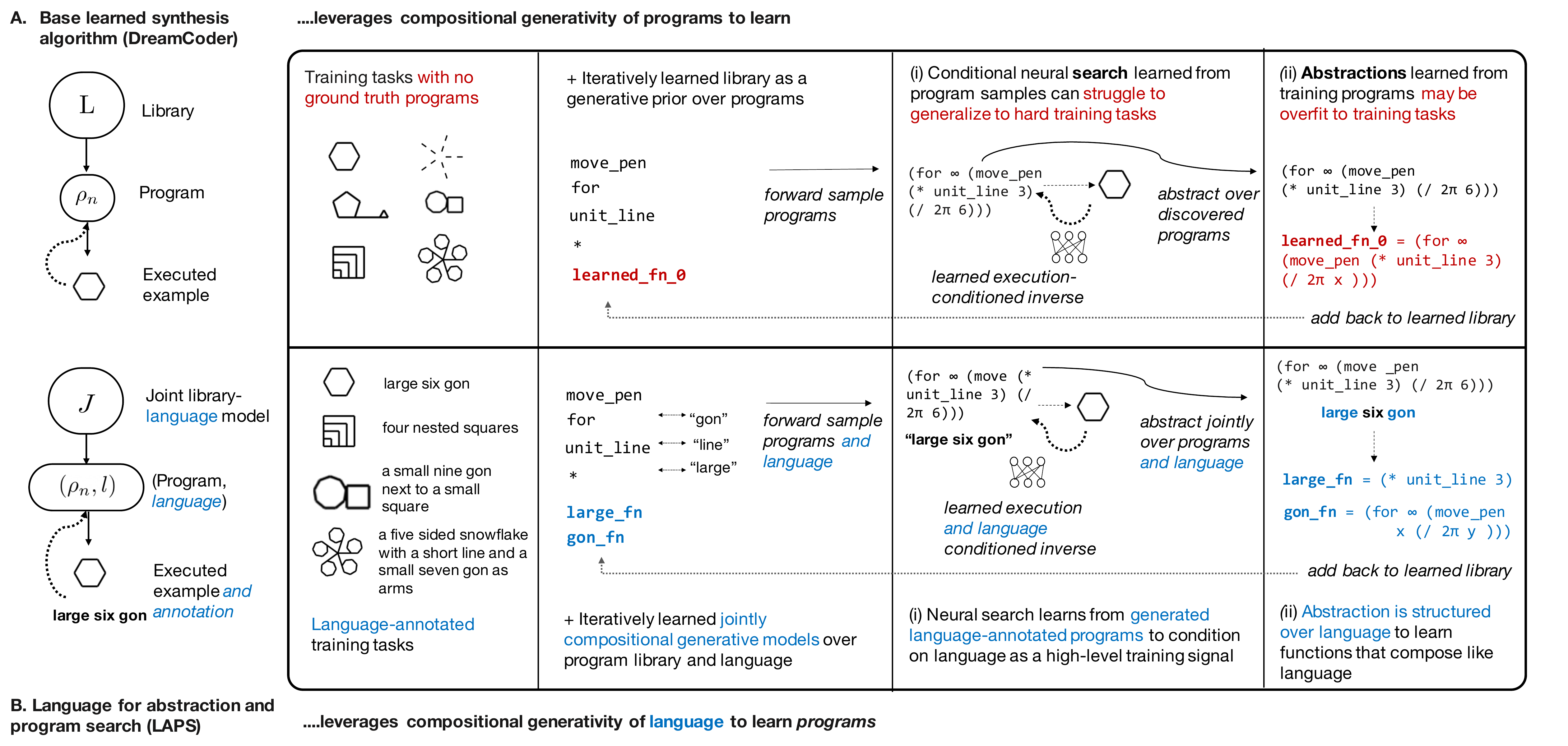}
  \caption{Our model, Language for Abstraction and Program Search (LAPS) integrates natural language into base learned synthesis algorithms formulated as hierarchical Bayesian inference (A, left) for jointly learning a \textbf{library} of program abstractions and a \textbf{neural search heuristic} for synthesis. We give an extended formulation (B, left) defined jointly over the program library and natural language descriptions of synthesis tasks, that can be used to incorporate natural language into both abstraction and search heuristic learning. When incorporated into a concrete learning algorithm, DreamCoder (A, right) we show that LAPS allows the model to leverage language richly during training to improve the generalization of both the learned neural search model and the learned library of program abstractions.}
\end{figure*}

Machine learning approaches based on program synthesis--the automatic inference of symbolic programs--can offer robustness, interpretability, verifiability, and strong generalization in few-shot learning settings \cite{appel2017position,lake2017building}. Many machine learning tasks can be formulated as program synthesis problems, including data manipulation \cite{delaware2015fiat, gulwani2017program}, semantic parsing \cite{artzi2013weakly, liang2016learning}, structured visual understanding \cite{johnson2017inferring, yi2018neural}, image generation \cite{ellis2017learning,ganin2018synthesizing}, and policy learning  \cite{fikes1971strips,cropper2015learning,silver2020few}.

This paper introduces \textbf{Language for Abstraction and Program Search (LAPS)}, a framework for improving the efficiency and generalizability of learned program synthesis models using natural language supervision. In LAPS, language guides learning of both \emph{libraries} of reusable program abstractions and \textit{heuristics} for searching in the space of programs. High-quality program libraries and search methods are the main ingredients of effective program synthesis approaches \cite{gulwani2017program}. Recent approaches to program synthesis have attempted to learn search models \cite{gulwani2015inductive,polozov2015flashmeta,balog2016deepcoder,devlin2017robustfill}, program libraries, or both jointly from data \cite{shin2019program,dumancicinventing,ellis2020dreamcoder,ellis2020dreamcoder-arxiv,lazaro2019beyond}, but even the current best learning approaches can be computationally inefficient (often requiring upwards of thousands of CPU hours to bootstrap learning) and do not always discover generalizable libraries or search strategies. 

LAPS builds on the intuition that natural language offers a powerful source of information for tackling both learning problems. Language simultaneously provides an efficient channel for communicating the structure of the search space
(an instruction like \textit{draw a large hexagon next to a small pentagon} decomposes a complex graphics task into high-level parts) and a lexicon that names important reusable concepts in a given domain (for instance, suggesting that a function to draw variable-sized \textit{polygons} might be useful for future graphics tasks). 
In this work we show how inducing \textit{jointly compositional generative models over natural language and programs} 
provides a strong scaffold for library and search model learning in a hierarchical program induction model. When integrated into a state-of-the-art learning algorithm, DreamCoder \cite{ellis2020dreamcoder,ellis2018learning}, our approach dramatically improves performance on three different synthesis domains: \textit{string editing, structured image generation} and \textit{scene understanding}. Compared to the base synthesis approach, LAPS solves and learns more quickly from synthesis tasks, and produces higher-quality libraries that improve generalization to downstream tasks \textit{without} natural language hints.


LAPS builds on several recent developments in (non-language-based) program synthesis, so we begin with a review of related work (\autoref{related-work}), then formalize the search and library learning problems (\autoref{problem-formulation}) and base synthesis algorithm (\autoref{section-dreamcoder}). We then describe how LAPS extends the base algorithm to include language in learning (\autoref{section-LAPS}) and conclude with empirical results (\autoref{experiments}).

\section{Related Work}\label{related-work}  
Our work draws on recent program synthesis approaches that \textit{learn to synthesize programs from examples} using neural models to guide search \cite{gulwani2015inductive,balog2016deepcoder,parisotto2016neuro,devlin2017robustfill,polosukhin2018neural,abolafia2018neural,nye2019learning,ellis2019write,si2019learning,ye2020benchmarking}; and \textit{learn libraries} of symbolic abstractions from a collection of related programs or tasks \cite{dechter2013bootstrap,zhang2017macro,shin2019program,dumancicinventing,ellis2018learning,ellis2020dreamcoder}. Our formulation builds on \textit{hierarchical Bayesian formulations of program learning} that frame both synthesis and library learning as probabilistic inference \cite{liang2010learning,lake2015human,ellis2020dreamcoder}.

Natural language has also been used to scaffold latent representation learning \cite{frome2013devise, jia2016data,andreas2017learning,ye2020optimal,goyal2020pixl2r,liang2020alice,mu2019shaping,luketina2019survey}, and as a high-level specification for program synthesis tasks \cite{ye2020benchmarking, nye2019learning,polosukhin2018neural,ye2020optimal,desai2016program,srivastava2017joint}. Here we present an approach that integrates language annotations in \textit{training} for learning a more generalizable library and program search model that can be used after training with no additional annotations for new tasks.

\section{Inductive synthesis and library learning}\label{problem-formulation}
Consider the problem of writing a graphics program to draw the large hexagon image in the left column of Fig.~\ref{banner}. This is an \textit{inductive program synthesis} problem: a task $t$ (like \emph{draw a large hexagon}) is \textit{specified with examples of what a program should do}, where each example is given as an input $x$ (in this case, the blank image canvas) and the desired output $y$ (the large hexagon image). A program $\program$ solves the task if it produces outputs that are consistent with the specification when executed -- that is, if evaluating $\program$ under an execution model $E$ yields $\eval{\program}_E(x) =y $.


Program synthesis begins with a \textbf{library}
$\library=\{l_0,..l_n\}$ containing the set of primitives that can be combined to produce solution programs, such as the (pseudo-code) primitive functions
in a simple graphics language:
$$\library = \small\texttt{move\_pen|unit\_line|for|*|$\pi$|$\infty$|0|1|2|} ...$$
which draw lines on a canvas parameterized by their length and angle.
Given a library, there is also the problem of \textbf{search}: effective program synthesis requires a search strategy $S$ that can be given a task specification (such as the image of a hexagon) and automatically discover a solution program like the one shown in Fig. \ref{banner}:
$$\small\texttt{(for $\infty$(move\_pen($*$ unit\_line 3)(/ 2$\pi$ 6))} $$ by searching over programs built from functions in $\library$. 

Both of these ingredients -- the \textbf{library} $\library$, and the \textbf{search strategy} $S$ -- can be made much more efficient if the synthesis engine will be expected to solve multiple related problems. 
In the graphics domain, for example, synthesis of the various images depicted in Fig. \ref{banner} is much more easily accomplished using a library like
$$\library= \small\texttt{polygon|large\_line|small\_line} ...$$
in which the original hexagon task can be expressed as 
$$\small\texttt{polygon(6, large\_line)}$$
A good library already provides a foundation for efficient search by making solutions easier to express.
%
%
%
Even with such a library, search can be further guided by information about the prior structure of programs (for example, the fact that \texttt{polygon} is typically called with a \texttt{large\_line} or \texttt{small\_line} function as a second argument) and by information about the target task itself (for example, the fact that the target image contains six line segments).
Thus, one way to describe an effective search strategy $S$ is via a \textit{prior} over programs $\probability[\program | \library]$ in the library and a \textit{conditional} inference model for inferring $\probability[\program | t, \library]$, the distribution over programs likely intended by the observed task examples $t$.

The foregoing discussion lays out the basic ingredients of a 
hierarchical Bayesian formulation of program synthesis (used in learning algorithms like \cite{ellis2020dreamcoder,lake2015human,dechter2013bootstrap}; see the graphical model in Fig.~\ref{banner}A, left) for jointly learning a library and conditional search model from a dataset $T$ of synthesis tasks.
We denote a prior over programs as $\probability[\program | \library, \libparams]$, on a library $\library$ with parameters $\libparams$. Given the observed tasks, we define the likelihood of the latent library and parameters as:
\begin{equation}\label{eq:joint-base}
      \Phi(\library,\libparams)=\probability[\library,\libparams]\prod_{t\in T} \sum_\program \probability[t|\program]\probability[\program|\library,\libparams]
\end{equation}
where $\probability[\library,\libparams]$ is a prior over all possible libraries and parameterizations, and $\probability[t|\program]$ is the likelihood that each inductive task $t$ is consistent with a program $\program$ (for our purposes, $\probability[t|\program] = 1$ if the program produces the desired output examples and $0$ otherwise.) Learning in this model means estimating the optimal library and its parameters 

\begin{equation}\label{optimal-library}
  \library^*, \theta_{\library^*}  = \argmax_{\library, \theta_{\library}} \Phi(\library,\libparams)
\end{equation} 

along with a conditional model $\probability[\program | t, \library^*]$ that can infer programs for new tasks.

This formulation also foreshadows a straightforward way in which linguistic \textit{descriptions} of tasks (like those in the first column of Fig. \ref{banner}) could be integrated into learning: we could simply extend the conditional model as $\probability[\program | t, \desc, \library^*]$ to include a task's description $\desc$. We come back to this (and describe a more complete integration) in our approach, but first describe a concrete implementation of Eq. \ref{optimal-library} on which we can realize the language-enriched model.

\section{Base learning algorithm: DreamCoder}\label{section-dreamcoder}
The LAPS framework we describe in this paper is a general one for extending Bayesian models of program learning like the one in Eq. \ref{optimal-library} to incorporate information from language. For concreteness, however, our presentation and experiments build on the specific DreamCoder algorithm of \citet{ellis2020dreamcoder}, which we briefly review here.
We choose DreamCoder because it exposes a modular implementation of the library and search learning problems in Eq. \ref{optimal-library} and has previously demonstrated state-of-the-art performance across a variety of synthesis domains \cite{ellis2020dreamcoder,ellis2020dreamcoder-arxiv}. 

DreamCoder is initialized with a base library $\library_0$ of starting primitives and a dataset of training tasks $T$. It returns a \textit{learned} final library $\library_f$ augmented with program abstractions and a learned neural search model $Q(\program | t, \library)$ that predicts high probability programs conditioned on the task examples. 
Learning is iterative: DreamCoder alternately searches for solution programs to the training tasks (given a current library $\library_i$ and search model $Q_i$) and updates the library and search model based on new solved tasks. We give details on each component below. 

\subsection{Program prior}\label{program-prior}
DreamCoder defines the prior over programs as a probabilistic context free grammar (PFCG; \citealt{johnson1998pcfg}) for programs generated as productions from a library $\library$ of functions $l \in \library$ \footnote{In addition to initial and learned functions, \citet{ellis2020dreamcoder} define $\library$ to also include any initial literals and a rule for generating variables, such that programs can be completely generated as productions from the PCFG. We use the same formulation.}. Formally, DreamCoder assigns a real-valued weight ${\libparams}_i$ to each library function, which when normalized yields a production probability $\probability[l | \library, \libparams]$. The prior probability of a program $\program$ is given by 
\begin{equation}\label{dreamcoder-prior}
\probability[\program | \library, \libparams] = \prod_{l\in \program}\probability[l | \library, \libparams]
\end{equation}
the weighted product of probabilities of all of its constituent library functions. As all $\probability[l | \library, \libparams] < 1$, this is equivalent to a \textit{description length} prior over programs: longer programs (with more constitutent elements) will have lower prior probability under Eq. \ref{dreamcoder-prior} since $\probability[l | \library, \libparams]$ monotonically decreases as $|\program| = |\{\l \in \program\}|$ increases.

\subsection{Amortized conditional inference} \label{amortized}
To identify programs that solve tasks $t$ while obtaining high probability under $\probability[\program | \mathcal{L}, \libparams]$,
DreamCoder trains a neural search heuristic $Q_i(\program | t, \library_i)$ at each iteration $i$ to approximate the inverse conditional model. The heuristic uses a neural model trained to predict programs written in the current library $\library_i$ according to the posterior: 
\begin{equation}\label{dreamcoder-posterior}
    Q_i(\program|t, \library_i) \approx
    \probability[\program|t,(\library_i,{\libparams}_i)]\propto\probability[t|\program]\probability[\program|(\library_i,{\libparams}_i)]
\end{equation}
conditioned on an encoding of the training examples (e.g. an embedding of the image in the task specification). This model is trained in the distant supervision setting (which begins with no supervised program data) by leveraging the forward generative model: sampling programs from the prior, executing them to produce observed tasks, and then minimizing $Q(\program|t, \library)$ in Eq. \ref{dreamcoder-posterior} on the sampled programs, conditioned on their executions. This generative training procedure is generally applicable to any neural implementation of $Q(\program | t, \library)$.  (But see \citet{ellis2020dreamcoder} and our supplementary material for additional details on the model architecture, which we reimplement in our experiments.)

\subsection{Abstraction learning as program compression (maximizing the likelihood of programs)} 
The DreamCoder algorithm also iteratively updates the library ($\library_i, \theta_{\library_i}$) to approximately optimize Eq. \ref{optimal-library} (finding $\library^*, \libparams^*$ which maximize the likelihood of  the inferred latent programs). \citet{ellis2020dreamcoder} leverage equivalence to a \textit{compression} problem defined over programs and the library. As discussed in \ref{program-prior}, the PCFG program prior is equivalent to a description length prior over programs. \citet{ellis2020dreamcoder} place an additional Dirichlet prior over the library description length:
\begin{equation}\probability\left[ \mathcal L\right]\propto\exp \left(-\lambda\sum_{\program\in \library}\text{size}(\program) \right)
\end{equation}
Estimating the optimal library then becomes the problem of inferring new library abstractions which can jointly compress the latent training programs (rewritten under the new library $\library_{i+1}$) and the description length $|\library_{i+1}|$ of the updated library (to optimize for shared abstractions across programs). This objective would still require inference over all possible ways of refactoring the latent programs under the updated library. \citet{ellis2020dreamcoder} approximate this by only considering candidate abstractions and program refactorings that can be found via an efficient lambda-abstraction algorithm. As an example, this could refactor the large hexagon program
$$\small\texttt{(for $\infty$(move\_pen($*$ unit\_line 3)(/ 2$\pi$ 6))}$$
to expose a candidate abstraction like
$$\small\texttt{$\lambda$x.(for $\infty$(move\_pen($*$ unit\_line 3)(/ 2$\pi$ x))}$$ while also rewriting the original program using this abstraction. Notably, this fragment -- which draws polygons with lines of length 3 for sides -- is not the most intuitively generalizable for the graphics domain. A programmer with more domain-specific prior knowledge would probably prefer an abstraction like
$$\small\texttt{$\lambda$xy.(for $\infty$(move\_pen($*$ unit\_line y)(/ 2$\pi$ x))}$$
which additionally parameterizes the polygon by the length of its sides, and is semantically equivalent to the high-level $\texttt{polygon\_fn}$ described in the problem setup in Sec.~\ref{problem-formulation}. However, learning abstractions by compressing the library and current solved training tasks may actually disfavor this more intuitively generalizable (but less compressive) candidate. Our second key goal in introducing language will be to leverage it as an additional source of prior knowledge to improve abstraction generalization.

\section{Our Approach: Language for Abstraction and Program Search}\label{section-LAPS}
Our work considers how the general learning problem -- jointly learning the library $ \library$ which defines the prior over programs and the conditional search strategy $S$ which inverts from tasks to programs -- can be enriched in the \textit{language-annotated} setting. Here, at least a subset of the training tasks are additionally annotated with a natural language description $\desc$ (such as the natural language description \textit{large six gon} for the large hexagon drawing task in Fig. \ref{banner}B). Language offers a more direct source of information for discovering a library like the one in our setup,
$$\library = \small\texttt{polygon|large\_line|small\_line} ...$$
if we leverage the expectation that generalizable abstractions (like a candidate \texttt{polygon} function) should correspond systematically to named fragments in natural language (like the token \textit{gon}).

Language can also be leveraged by the conditional search model: learning systematic correspondences between language and programs from descriptions like \textit{large six gon} should inform search on new tasks
(like the one described as a \textit{small nine gon next to a small square} in  Fig. \ref{banner}B) on the basis of shared language (like \textit{gon}).

Our approach, \textbf{LAPS (Language for Abstraction and Program Search)} formalizes these intuitions by extending the hierarchical Bayesian problem formulation over \textit{programs} given in \autoref{problem-formulation} to additionally generate \textit{natural language} task descriptions (see graphical model in Fig \ref{banner}B, left). In particular, we assume the existence of a \textit{jointly} generative model $J(\program, \desc)$ over latent programs that solve tasks, and corresponding natural language descriptions. We rewrite the original prior over programs $\probability[\program | \library, \libparams]$ defined on a library $\library$ to a \textit{joint} prior $\probability[\program, \desc | \joint, \jointparams]$, and extend the distribution in Eq. \ref{eq:joint-base} over the latent \textit{joint} model $\joint$ with parameters $\jointparams$, written as
\begin{equation}
      \label{eq:joint-extended}
      \Phi(\joint,\jointparams)=\probability[\joint,\jointparams]\prod_{t\in T} \sum_\program \probability[t|\program]\probability[\program, \desc|\joint,\jointparams]
\end{equation} 
Learning in the language-augmented setting now involves estimating the optimal joint model and its parameters

\begin{equation}\label{optimal-joint}
  \joint^*, \optimaljointparams  = \argmax_{\joint, \jointparams} \Phi(\joint,\jointparams)
\end{equation} 

along with a language-conditioned model $\probability[\program | t, d, J^*]$ that can infer programs for new tasks based on both specification examples \textit{and task descriptions}.

In the remainder of this section we first describe a general joint model formulation that can be learned from language-annotated training tasks. We then show how the joint framework allows natural language to inform learning at both the abstraction and search level in a concrete example, using DreamCoder as the base hierarchical algorithm.

\subsection{Joint prior over programs and language}
\paragraph{Base prior}
We formulate our joint prior over language and programs as 
\begin{equation} \label{joint-prior}
\probability[\program, \desc] = \probability[\program | \library, \libparams] \probability[\desc | \program, \library]
\end{equation}
decomposed as the product of the original program prior defined on a program library $\probability[\program | \library, \libparams]$, and a learned program-to-natural-language ``translation" model $\translation(\desc | \rho, \library) \approx \probability[\desc | \program, \library]$ which describes how natural language descriptions are generated for latent programs (in our running example, this model would describe how the \textit{large six gon} description was generated conditioned on the program solution for that task.) This decomposition builds modularly on the original program prior defined only on the library $\library$. Learning $\translation(\desc | \rho, \library)$ formalizes the intuition that there should be a learnable relationship between language that describes tasks and latent programs that solve them.

$\translation(\desc | \rho, \library)$ can be implemented in many ways (e.g. \cite{wong2007learning,joshi1997tree,bahdanau2014neural,chen2018tree}), compatible with the vast literature on structured translation between languages, including natural languages and programming languages. 
%
Our experiments use the translation model popularly known as \emph{IBM Model 4} \citep{brown1993mathematics}, one of a class of well-studied Bayesian machine translation models \cite{gal2013systematic} which decompose $\translation(\desc | \rho, \library)$ into
\begin{equation} \label{decomposed-translation}
\translation(\desc | \rho, \library) \propto \prod_{w \in \desc, l \in \program} \probability_\translation[w | l]
\end{equation}

a product of learned token-level translation probabilities $\probability_\translation[w | l]$ between individual functions $l$ in a task's latent program $\program$ and words $w$ in the task description $\desc$. (See supplementary materials for model implementation and training details.) This token-level decomposition more directly captures the intuition in our setup: that abstractions in a programming library generally correspond systematically to individual names in natural language descriptions, and that the inverse conditional search can be guided based on a generally compositional relationship between program primitives and words. This formulation also allows these compositional relationships to be inferred from fewer observed examples than would be possible with other translation models with weaker inductive biases. However, Eq. \ref{joint-prior} should extend to include any similar translation model and need not include this stronger decomposition.
\vspace{-0.2cm}
\paragraph{Adding richer priors}
In LAPS, the joint model can also provide a controllable interface for incorporating additional prior knowledge about language into learning. Learned translation models are often fit to only maximize the likelihood of the observed language (here, with respect to inferred latent training programs). However, our formulation also supports $\translation(\desc | \rho, \library)$ enriched to include additional priors over language (such as speaker-specific language usage, or \textit{pragmatics} models that capture a speakers' other communicative goals \cite{grice1989studies,goodman2016pragmatic}.) 

In our experiments (\autoref{section-experiments}) we showcase this with results from an extended model incorporating an additional \textbf{mutual exclusivity} prior. Mutual exclusivity models the expectation that newly encountered words should correspond to different meanings than known ones. This prior has been shown to play an important role in language learning in cognitive science \cite{frank2009using,markman1988children}, and in machine learning models \cite{gandhi2019mutual}. 

In the synthesis setting, mutual exclusivity can capture the expectation that ``new" words (which appear in descriptions of currently unsolved tasks) are more likely to correspond to different program components than those used in solved training tasks (and for which there would otherwise be no signal to learn a translation model in the distant setting). Our extended model incorporates this prior by updating Eq. \ref{decomposed-translation} to distinguish between $W_{known}$ (words that appear in solved training tasks with latent programs) and $W_{new}$ (newly encountered words) as
\begin{equation} \label{decomposed-translation-me}
\begin{split}
\translation_{ME}(\desc | \program, \library) \propto \prod_{w \in d, l \in \program} (\mathbbm{1}[w \in W_{known}] \probability_\translation[w | l]) \\
(\mathbbm{1}[w \in W_{new}] \probability[l | \library, \libparams]^{-1}])
\end{split}
\end{equation}
where new words are modeled as \textit{inversely} related to primitives under the program prior (fit to previously solved tasks) -- modeling the expectation that new words more likely relate to less-used program components than those used so far.

\subsection{Integrating the joint model into amortized conditional search}
The joint model allows LAPS to incorporate natural language into the learned conditional search model over programs. In place of the original neural amortized model in the base algorithm (\autoref{amortized}), we train an extended, language-conditioned model $Q_i(\program | t, \desc, J_i)$ at each iteration to predict programs according to:
\begin{equation}
\begin{split}
    Q(\program | t, \desc, J_i) &\approx \probability[\program|t, \desc, J,\jointparams] \\ & \propto\probability[t|\program]\probability[\program, \desc|\joint,\jointparams] \\
    &\propto\probability[t|\program]\probability[\desc|\program]\probability[\program |\library,\libparams]\\ &
    \approx \probability[t|\program]\translation(\desc|\program, \library)\probability[\program |\library,\libparams]
\end{split}
\end{equation}
which amortizes program inference under our joint model formulation. Importantly, we can train this neural model using samples from the \textit{joint} generative model, consisting of sampled programs \textit{and corresponding generated language}. As with the original learning setting, this sample-based training allows LAPS to learn a generalizable, language-conditioned neural search heuristic, capable of leveraging compositional patterns in natural language, from very few examples in the distant supervision setting. We can also now see the benefits of richer language-specific priors (such as mutual exclusivity): the neural model trained to amortize inference from the joint generative model can also approximate the mutual exclusivity bias, enabling better exploration and generalization in the presence of new words.

\subsection{Abstraction learning as joint model compression}
The extended joint model objective in Eq. \ref{optimal-library} and \ref{optimal-joint} also allows LAPS to incorporate natural language into \textit{abstraction learning}. Extending the compression-based abstraction objective in the base algorithm -- which optimized for libraries that maximally compress the latent training programs and library -- requires defining a prior over the language-program \textit{translation} model $\translation$ in terms of the optimal program library.

We place a prior over $\translation$ defined on a program library $\library$ and a natural language token vocabulary $W$ as 
\begin{equation} \label{translation-prior}
    \probability[\translation | \library] \propto  \sum_{l \in \library, w \in W}-I(\probability_\translation[w|l])
\end{equation}
where $-I(\probability_\translation[w|l]) = -\log(\probability_\translation[w|l])$. This models the intuition that a good library contains program abstractions which correspond well to individual language tokens, and reduce entropy in the compositional translation model. Defining the prior compositionally also allows the algorithm to maintain the desirably property from  \cite{ellis2020dreamcoder}, in which the joint likelihood can be efficiently re-approximated with respect to individual candidate program abstractions based on their constituent subcomponents $l$ and corresponding translation distributions $\probability_\translation[w|l]$ under the current translation model. As in the base synthesis algorithm, we fully re-estimate a new translation model at each iteration $\translation_{i+1}(\desc |\program_{i+1}, \library_{i+1})$ to fit the updated library and refactored programs. See the supplement for extended details.

Taken together, Alg. \ref{algorithm-1} summarizes the concrete algorithm using LAPS to incorporate language into  \cite{ellis2020dreamcoder}.

\begin{algorithm}[t]
\caption{ }
\label{algorithm-1}
\begin{algorithmic}
\STATE \textbf{Input:} Initial library $\library_0$, annotated training tasks $(T, D)$
\STATE Initialize $J \gets$ uniform; training task solutions \textbf{p} $\gets \{\}$
\FOR{i $\leq f$} 
    \STATE $Q_i(\program | t, \desc)$ $\gets $ Train on (\textbf{p}, T, $\desc$) and samples $\sim J$ 
    \STATE \textbf{$p$} $\gets $ programs from search amortized with $Q_i$
    \STATE $\library_i \gets$ abstractions optimized over (\textbf{p}, J)
    \STATE \textbf{$p$} $\gets $ programs rewritten using abstractions from $L_i$
    \STATE $J_i \gets$ Fit $\libparams$ and $\translation(\desc | \rho)$ to (\textbf{p}, $\desc$)
\ENDFOR
\STATE \textbf{Return} $Q_f, \library_f$
\end{algorithmic}

\end{algorithm}

\section{Experiments}\label{experiments}
\begin{figure*}[h!]
\centering
  \includegraphics[width=0.93\textwidth]{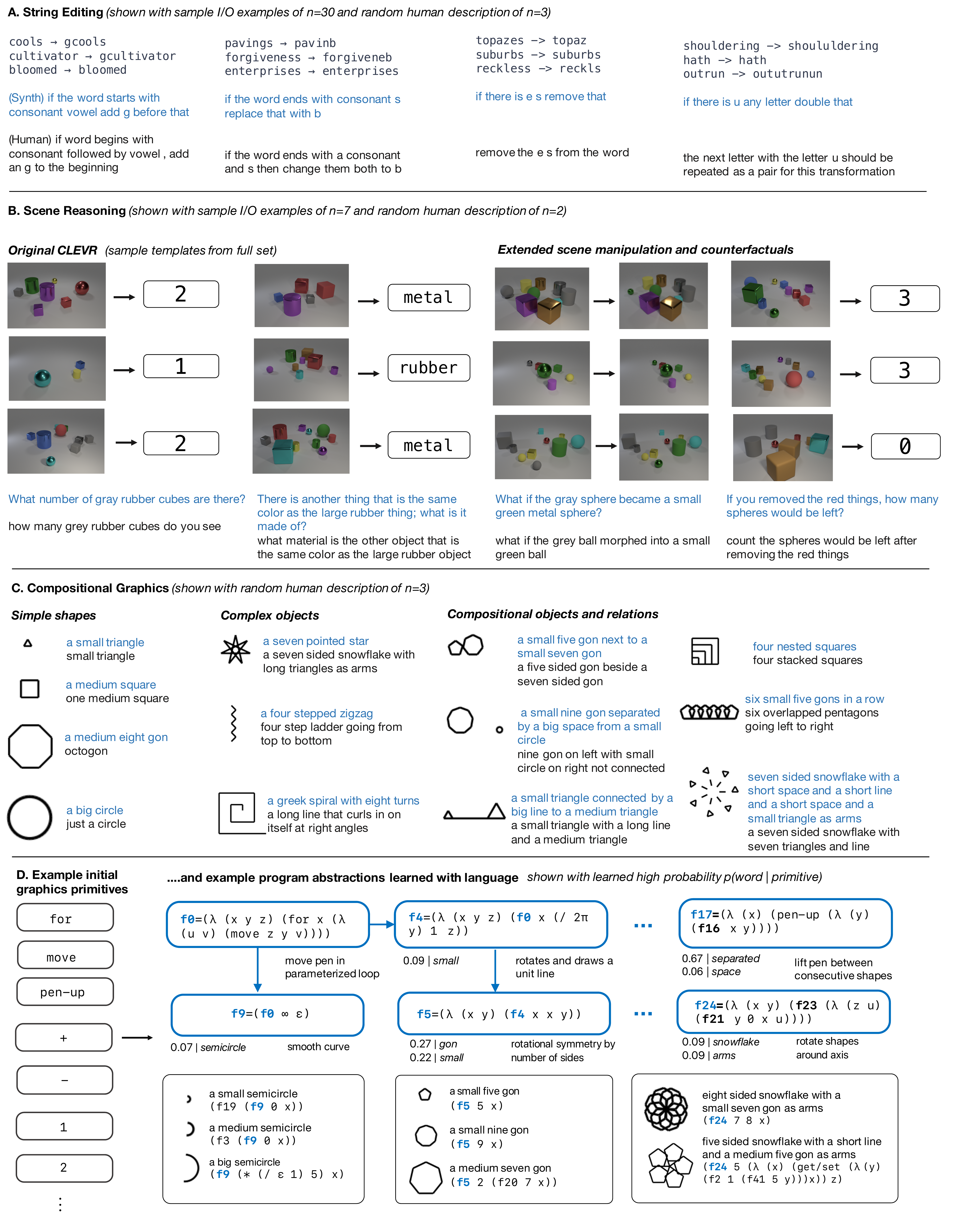}
      \caption{\textbf{(A, B, C)} Example tasks from all three synthesis domains shown with synthetic and sample human language annotations. Inductive synthesis domains are shown with a random subset (n=3) of the paired input/output examples. Human language annotations are also randomly sampled (all domains were annotated by multiple people for a broader range of language.) \textbf{(D)} Representative \textit{initial program primitives} and \textit{library abstractions} learned with LAPS for the graphics domain. Shown with example tasks solved with synthesized programs containing the learned abstractions and high probability natural language learned from the joint model. }\label{domains}
\end{figure*}

We demonstrate LAPS on three different domains: \textit{string editing, compositional graphics drawing}, and \textit{scene reasoning}, which we choose to represent a diverse range of tasks and accompanying language (Fig. \ref{domains}). In all three domains, we find that compared to the base synthesizer, LAPS learns and solves heldout synthesis problems faster (Table \ref{quant-results}, Sec. 1-2), and produces higher-quality libraries that improve generalization even when natural language hints are \textit{not} available after training (Table \ref{quant-results}, Sec. 3). 

Below we summarize each domain. We then discuss results showing that LAPS is effective because of how the hierarchical model incorporates language during learning: we find that (1) \textit{LAPS searches more effectively} during training, enabling it to solve and learn from more diverse training tasks than the baseline model; (2) \textit{LAPS abstracts more effectively during training}, adding in more generalizable library routines as it learns; and (3) LAPS \textit{can} use language during testing if it is available, as an important additional source of high-level information during synthesis.

\subsection{Domains} \label{section-domains}
\label{section-experiments}
All three domains consist of a dataset of inductive synthesis \textit{tasks} $t$ specified as input/output examples; procedurally generated \textit{synthetic language annotations}; and \textit{human language annotations} sourced from Mechanical Turk. We use synthetic language as our primary evaluation benchmark: we are interested in a controlled probe of learning when words are systematically reused and composed, but refer to more abstract concepts than in the initial base programming language. However, we also use human language to evaluate the practicality of our approach in real-world settings.\textit{ Additional information for all domains is in the supplement}.

\textbf{String editing:} structured string transformation problems taken from \cite{andreas2017learning} (n=1000 train; n=500 test). Tasks consist of input dictionary strings transformed using randomly sampled regular expression transducer (30 I/O examples per task). We choose this domain to demonstrate LAPS on an important classic synthesis domain \cite{lau1998programming}. The dataset of \citet{andreas2017learning} contains human annotations; synthetic language  annotations are generated over the ground-truth regexes using templates based on the original human annotations. We initialize synthesizers with functional programming primitives (\textit{map, fold, cons, car, cdr, length, index}) and character constants (following the simpler text editing domain in the baseline paper \cite{ellis2020dreamcoder}). The neural search model encodes the I/O task examples as character arrays with a bidirectional GRU.

\textbf{Compositional graphics}: inverse graphics problems (n=200 train; n=111 test) where each task is specified by an image and solved by synthesizing a program in LOGO Turtle graphics \cite{abelson1986turtle}. This is inspired by the graphics domain in \cite{ellis2020dreamcoder} but re-designed to be more challenging (ground-truth programs are much longer on average in the base programming language) and explicitly compositional. Synthetic language annotations are generated with high-level templates over the objects and relations in each task; human annotations are sourced as image descriptions from MTurk. We initialize synthesizers with the graphics primitives in \cite{ellis2020dreamcoder}. The neural model encodes image examples with a CNN.

\textbf{Structured scene reasoning:} inductive scene reasoning tasks (n= 212 train; n=115 test) where each synthesis problem is specified by a structured input scene, and outputs can be a number (\textit{how many red rubber things are there?}), a boolean value (\textit{are there more blue things than green?}), or another scene (\textit{what if all of the red things turned blue?}). This domain is modeled on 
CLEVR 
\cite{johnson2017clevr} but designed to support inductive 
synthesis tasks 
specified over the symbolic scene representations (an array of objects represented as dictionaries of attributes) from the original CLEVR task generator in \citet{johnson2017clevr}. 
We also add 
new tasks that require \textit{generating} or \textit{imagining} latent scenes (\textit{how many metal things would be left if all the blue cylinders were removed?}), 
which are not solvable in the original high-level DSL hand-designed for \citet{johnson2017inferring} (and used in synthesis-based approaches like \citet{yi2018neural}). We
include these to demonstrate a key feature of our approach: the ability to \textit{learn} generalizable libraries from a basic but expressive set of primitives, rather than restricting the program space pre-emptively with a hand-designed language. We use synthetic language annotations from the original templates in \cite{johnson2017clevr} (and templates written in the same style for the extended tasks); human annotations are sourced from annotators shown the same tasks. We initialize synthesizers with functional programming primitives similar to the string-editing domain, with domain-specific query functions and constants (\textit{get\_color(x); get\_shape(x); blue; cube}). The neural model encodes the task examples as flattened arrays of object attributes using a bidirectional GRU.
\subsection{Results}\label{results}

\begin{table*}[]
\caption{\% held-out test-tasks solved. To compare robustness, we run random seed replications in the graphics domain for the synthetic language dataset. \textit{Best} reports the best model across replications; \textit{Mean} averages across replications. }
  \label{quant-results}
\resizebox{\textwidth}{!}{%
\begin{tabular}{@{}p{0.3\linewidth}lccccc@{}}
\toprule
Language & Model                                                & \multicolumn{1}{l}{Strings (n$_{test}$  = 500)}   & \multicolumn{2}{c}{Graphics (n$_{test}$ = 111)} & \multicolumn{2}{c}{Scenes (n$_{test}$  = 115)} \\ \midrule
                          &                                     & \% Solved          & \% Solved (Best) & \% Solved (Mean)   & \% Solved (Curric.) & \% Solved (Mean.)         \\ \midrule
Synth train/test                   & DreamCoder (no language)             & 33.4           & 49.55            & 42. 64           & 67.80   & 73.9                     \\
Synth train/test                & Multimodal (no generative translation model)                    & 46.00         & 26.12            & 23.20             & 76.50   & 49.5                             \\
\midrule
Synth train/test                       & LAPS in  neural search     & 52.20        & 92.79            & 52.93              & 95.6    & 88.1                        \\
Synth train/test                            & LAPS + mutual exclusivity        & \textbf{57.00}        & 86.49            & 80.18      & \textbf{96.5}   & 82.3                     \\
Synth train/test                            & LAPS + ME + language-program compression & 54.60     &\textbf{98.19} & \textbf{81.98}            &  95.6     & \textbf{95.9}                              \\ \midrule        
Synth train/human test & LAPS + ME + language-program compression & 54.60  & 89.20            & --                       & 97.4     & --                            \\
Human train/human test & LAPS + ME + language-program compression  & 48.60            & 58.55            & --            & 95.6   & --               \\ \midrule
\textbf{No language at test} &       & \multicolumn{1}{l}{} & \multicolumn{1}{l}{} & \multicolumn{1}{l}{}   & \multicolumn{1}{l}{}                  \\ \midrule
No language on train/test & Original DSL;  Enumerative              & 0.06  & 0.00             & --                 & 27.8  & --               \\
No language on train/test & DreamCoder (best library): Enumerative   & 27.2  & 41.44            & --                   & 53.6   & --           \\
No lang at test & LAPS (best library): Enumerative           & 33.2   & 62.16            & --                   & 93.04   & --            \\
No lang at test & LAPS (best library): example-only neural synthesis       & \textbf{52.4}       & \textbf{91.0}             & --               & 95.6   & --         \\ \bottomrule
\end{tabular}%
}
\end{table*}
\begin{figure*}[h!]
  \centering
   \vspace*{-10pt} 
       
    \includegraphics[width=\textwidth]{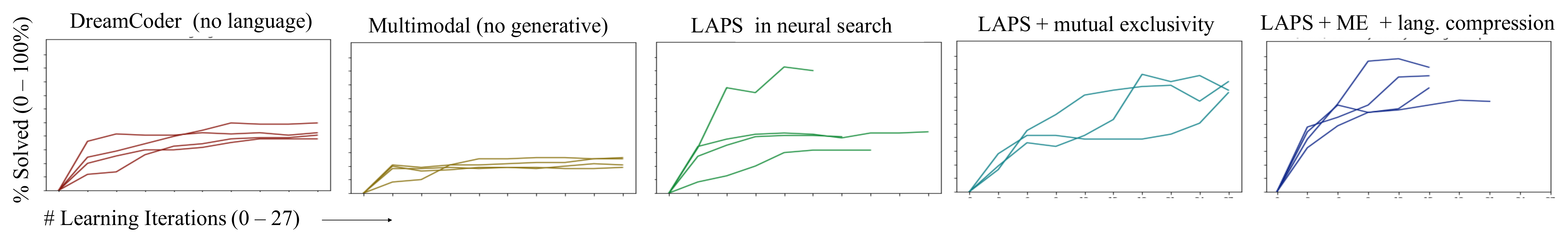}
   \vspace*{-20pt} 
  \caption{Learning curves comparing baselines and LAPS models in Table \ref{quant-results}, showing \% heldout tasks solved on the graphics domain over random training task orderings. (\textit{Mean} results in Table \ref{quant-results} shows average test-time performance from the trained model replications.)}
  \label{graphics-curves}   
\end{figure*}

On all three domains, we compare our model against the baseline synthesizer (Table \ref{quant-results}, \textbf{DreamCoder, no language}); a multimodal baseline (Table \ref{quant-results},\textbf{ multimodal, no generative model}) that trains a neural model directly on solved training tasks (similar to neural synthesis models like DeepCoder \cite{devlin2017robustfill} but augmented to condition on language); and ablated LAPS variants (Table \ref{quant-results}; \textbf{LAPS} rows) to evaluate the additive contributions of the individual learning components. We compare all models using a matched search budget per task and number of training iterations overall, determined using a hyperparameter search with the baseline. The supplement contains full details (and code) to replicate all experiments; and additional qualitative results. 

We find that:

(1) \textit{LAPS searches more effectively during training}, enabling it to solve and learn from more training tasks than the baseline synthesizer.  Under the hierarchical model formulation, search and abstraction are closely related: successfully solving tasks is the basis for abstraction learning.

Comparing the model \textit{learning trajectories} (Fig. \ref{graphics-curves}) on training tasks shows that the LAPS models consistently search more effectively during training: at each iteration they solve more tasks within a given time budget. Fig. \ref{graphics-curves} also highlights that LAPS models improve training \textit{robustness} in the distant learning setting: as in the baseline paper \cite{ellis2020dreamcoder}, we find the baseline model learning to be highly variable without a training curriculum (compare training curves from Fig. \ref{graphics-curves} with different random seed replications; and the \textit{best} vs. \textit{mean} performance, Table \ref{quant-results}.) Comparing the LAPS ablations also suggests that linguistic priors (like \textit{mutual exclusivity}) can indeed be practically useful here during learning (Table \ref{quant-results}, compare \textit{LAPS with ME and without}).

What if we do use a curriculum? In the scene reasoning 
domain (where previous approaches (e.g. \citealt{mao2019neuro}) have argued for a curriculum), we also test a simple curriculum 
by ordering tasks according to their natural language token length (which can be evaluated without ground truth programs). Table 1 shows that our model is still more effective, and that non-curriculum performance is in fact comparable to curriculum performance.

(2)  \textit{LAPS abstracts more effectively during training}, adding in more generalizable library routines as it learns. The variability across training replications in the baselines also highlights a challenge for abstraction learning: not all shared subroutines encountered in training generalize well to new tasks. Adding poor abstractions can actually be detrimental: they increase the combinatorial search space. We find that our approach produces higher-quality libraries after training: Table \ref{quant-results} (\textbf{no language at test time} section) shows that we 
consistently improve performance in a head-to-head comparison using enumerative search from the library priors alone -- in some domains, enumerative search with our model’s library outperforms \textit{neurally guided search} from the baseline model.  We
also find the learned library is effective for neurally-guided 
synthesis when no language hints are available after training (Table \ref{quant-results},\textbf{ no language at test, example-guided synthesis}), showing that LAPS incorporates language to learn a more effective library overall, which generalizes to the non-language setting. See supplement for example learned abstractions from $\library_f$.

(3) \textit{LAPS can use language during testing if it is available, though it doesn't need to for competitive performance}. Clearly, language can provide a useful source of high-level information if it is available for new tasks. Our approach produces a neural synthesizer pre-trained to condition on language where available. Results on all three domains show that the model can use it to achieve additional performance gains (Table \ref{quant-results}, see  \textit{language at test} rows). We also find that the models trained on synthetic annotations generalize effectively to natural human language at test (Table \ref{quant-results}, \textit{synth train, human test}), suggesting that even if human annotation is too costly, in many cases hand-writing natural language templates to accompany a few ground-truth programs is likely sufficient (and easier than hand designing a full DSL).

\section{Conclusion}
We presented \textbf{Language for Abstraction and Program Search} (LAPS). LAPS builds on hierarchical Bayesian models of program learning: we offer a general framework for introducing \textit{jointly generative} models over programs and language into learned synthesis. Going forwards, an important avenue for future work will be exploring different concrete implementations of the base algorithm and translation model which relates programs to language. A promising future direction could leverage recent structured, neural joint models that can \textit{learn} the compositional units of language, and incorporate pre-trained language representations  \cite{joshi1997tree, wiseman2018learning, kim2019compound}.

The hierarchical Bayesian framing also draws connections to computational cognitive models which model \textit{human conceptual representations and learning} \cite{goodman2014concepts,fodor1975language,rule2020child} as inference over program-like representations. Future \textit{human} experiments could explore LAPS as a cognitive model, combining paradigms for studying language learning with those for studying non-linguistic abstraction and search (e.g. \citealt{smith2003complex,hawkins2019emergence,lake2015human,lake2019human,tian2020learning}).

\let\thefootnote\relax\footnotetext{\textbf{Acknowledgements}: Many thanks to  M. Nye,  J. Mu,  A. Marzoev, J. Fan, R. Hawkins, R. Levy, L. Schulz and our anonymous reviewers for invaluable feedback. Supported by grants from the Air Force Office of Scientific Research, the NSF under Grant No. 1918839 and NSF-funded Center for Brains, Minds, and Machines, the MIT-IBM Watson AI Lab, Google, Microsoft and Amazon.}

\bibliography{icml_2021_paper}
\bibliographystyle{icml2021}

\end{document}


\twocolumn[
\icmltitle{Supplemental: Leveraging Language to Learn Program Search Heuristics and Abstractions}
This contains the supplemental appendix to the 2021 ICML paper. It is organized sequentially in reference to the main text; S\{N\} refers back to section N in the main text. 

A complete release of code for our implementation, including command line scripts to replicate the experiments in the paper and links to the datasets, can be found at: \texttt{https://bit.ly/3g9361W}.
\vskip 0.3in
]

\section*{S4. Base learning algorithm: DreamCoder}\label{supplemental-dreamcoder}
The LAPS framework described in the main paper (Sec. 5) is a general one  for extending Bayesian models of program learning to incorporate information from natural language (see \cite{liang2010learning,lake2015human, dechter2013bootstrap,lake2013one}). Our concrete implementation and experiments use the DreamCoder approach of \cite{ellis2020dreamcoder, ellis2018learning} as the base synthesis algorithm, which implements the hierarchical Bayesian formulation of program learning. It defines a modular interface with two primary learning components: a learned \textit{conditional inference} model for search (as a neural search heuristic); and a learned \textit{abstraction} algorithm for updating the program prior (based on program refactoring and compression) \cite{ellis2020dreamcoder}. Each of these learning components has been additionally implemented in other work (such as \cite{devlin2017robustfill,polosukhin2018neural,nye2019learning,parisotto2016neuro,balog2016deepcoder} for neurally guided synthesis, and \cite{dechter2013bootstrap,zhang2017macro,shin2019program,artzi2014learning,dumancicinventing} for program abstraction learning). 

This supplementary section provides theoretical and implementation details on the DreamCoder algorithm we use in our experiments (summarized in Sec. 4). We match our implementation as closely as possible to the original work for comparison with published baselines. We provide key details relevant to the language-guided extension, but strongly recommend the original works which introduce the DreamCoder algorithm  \cite{ellis2020dreamcoder, ellis2018learning} for further reference.


\subsection*{S4.1 Program prior and MDL equivalence}
Hierarchical Bayesian program learning formulations require a prior over expressible programs. DreamCoder is learned iteratively: it is initialized with a base library $\library_0$ and returns a library $\library_f$ containing program abstractions learned from solving training tasks. Therefore, DreamCoder defines its program prior with respect to the current  library $\library_i$ maintained at each iteration. This is parameterized as a simple PCFG $\probability[\rho | \library, \libparams]$ whose productions are of the form $l_i \to l_j \in \library$, each with a real-valued weight $\theta_{\library l}$, where the probability of a program $\rho$ is given by $\probability[\rho | \library, \libparams] = \prod_{l\in \rho}\probability[l | \library, \libparams]$ (Sec. 4.1).

Minor complexity arises in order to support typing \cite{pierce}: following \cite{ellis2018learning}, the library $\library_i$ is implemented as a set of polymorphically typed $\lambda$-calculus expressions. The only change this produces to the original prior definition is to restrict the set of possible productions under the PCFG: that is, permissible productions are of the form $l_i \to l_j \in \{\library |l_i \to l_j \mkern3mu \text{is well typed}\}$. The prior probabilities of programs are therefore calculated with respect to the set of well-typed productions. 

As discussed in the main paper, this prior definition is \textit{equivalent to a minimum description-length prior over programs} under ($\library, \libparams$) when all $\libparams < 1.0$, as the product of additional productions in an expression will strictly decrease as the number of productions in an expression increases.

\subsection*{S4.2 Amortized conditional inference}
\begin{figure} [h]
    \centering
    \includegraphics[width =0.45\textwidth]{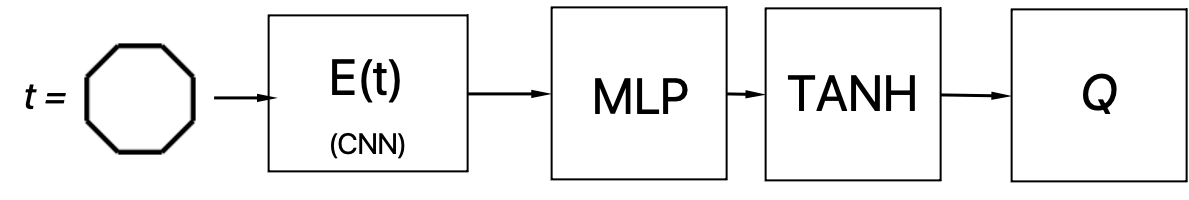}

    \caption{Architecture of the neural model $Q_i(\rho | t, \library_i)$. The model takes as input task examples $t$. These are encoded using a domain-specific encoder $E(t)$. Task encodings feed to an MLP and activation layer and output a tensor $Q$. This parameterizes a distribution over program bigrams in the final DSL, which defines a conditional distribution from which to enumerate programs during search.}
     \label{neural_model}
      
\end{figure}

To identify programs that solve tasks $t$ while obtaining high probability under $\probability[\rho | \mathcal{L}, \libparams]$,
DreamCoder trains a neural search heuristic $Q_i(\rho | t, \library_i)$ at each iteration $i$ to approximate the inverse model.

The training procedure in \cite{ellis2020dreamcoder} (summarized in Sec. 4.2) is a key contribution of the original work for learning in the distant supervision setting. The model is trained on samples from the generative prior (providing an endless training stream of random synthesis tasks); and this procedure should generalize immediately to any neural model for predicting programs conditioned on the task specification (e.g. \cite{devlin2017robustfill,polosukhin2018neural,nye2019learning,parisotto2016neuro,balog2016deepcoder}). The model is also supervised on any original training task examples and their program solutions discovered during learning.

In our experiments we use the baseline neural model architecture in \cite{ellis2020dreamcoder}. This is parameterized by two modular components:

\begin{enumerate}
    \item \textit{A domain-specific task encoder} $E(t)$. This encodes the task examples (e.g. \textit{images} in the graphics program domain, or input-output strings in the text editing domain) that are input to the neural model. This task encoder architecture is defined domain-specifically based on the form of the task examples (e.g. a CNN for the graphics domain). It outputs a fixed dimensional embedding for any given task as \textit{input} to the model. In our experiments this is a 64-dimensional embedding across all domains (See S6.1 for domain-specific architectures; and released code.)
    \item \textit{A conditional model over programs} $Q(\rho | E(t))$. This component receives the task encoding as input and outputs a distribution over programs. Following \cite{ellis2020dreamcoder}, this is a 2-layer fully-connected MLP (with 64 hidden units and a final tanh activation layer) that outputs a fixed-dimensional real-valued tensor encoding a distribution over programs in the library $\library$ as output. The real-valued tensor corresponds to weights over program primitives conditioned on their local context in the syntax tree of the program, consisting of the parent node in the syntax tree and which argument is being generated. This functions as a `bigram transition model' over trees that encodes the likelihood of transitions from one primitive to the next. $Q$ returns this as a $(|\library|+1)\times (|\library|+2)\times A$-dimensional tensor, where $A$ is the maximum arity of any primitive in the library. 
\end{enumerate}
This parameterization supports fast sampling of programs during conditional synthesis: the neural model runs once per task (to encode the task examples and produce the bigram transition model) and the resulting parameterization can then be used to sample programs during synthesis (e.g. by enumerating programs by expanding trees (as `bigrams' over parent and children primitives) ranked in order of their likelihood starting from the program root.)

Following \cite{ellis2020dreamcoder}, the neural model is trained to optimize the following MAP inference objective on the training tasks and the sampled tasks from the prior:
\begin{equation}
     \mathcal{L}_{\text{MAP}} = \expect_{\task\sim(\library,\libparams) }\left[ \log Q\left(\argmax_{\substack{\program}} \probability[\program |\task,\library,\libparams]\;\bigg\vert\; \task \right) \right]
\end{equation}

\subsection*{S4.3 Abstraction learning as program compression}
DreamCoder learns new abstractions to approximately optimize for Eq. 2 (main paper), which infers an optimal library and parameters with respect to the observed programs on the training tasks.

The DreamCoder abstraction algorithm is a primary contribution of the original work in \cite{ellis2020dreamcoder}, and is discussed extensively in \cite{ellis2020dreamcoder}. We therefore provide additional technical details here that are relevant to its integration with LAPS in our experiments, but strongly encourage referencing \cite{ellis2020dreamcoder} for the full implementation.

As discussed in \cite{ellis2020dreamcoder} and our main work, DreamCoder approaches abstraction using an equivalence between Eq. 3 and the \textit{minimum description length} of the \textit{prior} (as the description length of the library) and the \textit{programs} produced from the prior (under the PCFG definition of the prior). Therefore, in practice, inferring the optimal library is equivalent to inferring the library which maximally compresses the description length of the library and the description length of programs which explain the training tasks. In particular, DreamCoder optimizes the following compression objective with respect to the training tasks $T$ and the finite \textit{beam} $B_t$ of program solutions discovered for each training task during learning: 
\begin{align}
   \log \probability[\mathcal{L}] + \argmax_{\libparams}&\sum_{t\in T}\log \sum_{\program\in \mathcal{B}_t}\probability[t|\program]\max_{\program'\manyReduce\program}\probability[\program'|\mathcal{L},\libparams]\nonumber\\& \text{\phantom{tt}}+ \log \probability[\libparams|\mathcal{L}] 
    - \vert\libparams\vert_0 
\label{infiniteReFactorObjective}
\end{align}
The key aspect of this algorithm is that it considers abstractions which compress not only the programs as they are \textit{currently written}, but any semantically equivalent \textit{refactorings} of these programs. Specifically, as programs are written in a $\lambda$-calculus, \textit{refactoring} refers to any program which is equivalent up to $\beta$-reduction (i.e., function application/variable substitution~\cite{pierce}). A primary contribution of the original work in \cite{ellis2020dreamcoder} is an efficient algorithm for computing these refactorings that is unchanged when we integrate language; we refer to the original text for details.

In our work, the primary important aspect of this aspect is that  refactorings are defined compositionally over the existing program primitives. Specifically, refactorings can be efficiently calculated according to semantic equivalences in the the $\lambda$-calculus (namely, that function application and variable substitution guarantee that the resulting refactored programs are equivalent. \textit{Abstractions} created by variable substitution will always be composed of subcomponents from the initial library.) We take advantage of this compositionality when defining our joint abstraction algorithm over natural language. Defining an initial \textit{compositional} translation model between language and the program components ensures that we can approximate compression in the joint model after the programs are refactored, without needing to induce an entirely new translation model over language and the refactored programs.

\section*{S5. Our Approach: Language for Abstraction and Program Search}
This section now describes technical details for the concrete LAPS implementation in our reported experiments, which is defined over the DreamCoder implementation. We structure this section according to the parallel implementations in the base algorithm for clarity. However, except for the specifics of the joint-abstraction algorithm, the technical implementation of each component should extend directly to most other similar learned synthesis algorithms (e.g. the joint model implementation should be reusable in \textit{any} synthesis algorithm that uses an explicit symbolic library of primitives.)

\subsection*{S5.1 Joint prior over programs and language}
LAPS extends the prior $\probability[\rho]$ over programs under the library to a \textit{joint} prior $J(\program, \desc)$ over programs for a given task and their natural language descriptions $\desc$ (Sec. 5.1). We formulate this prior as 
$$J(\program, \desc) = \probability[\program | \library, \libparams] \probability[\desc | \program, \library]$$

the product of the original prior over programs $P[\rho | \library, \libparams]$ defined on the program library, and a \textit{program to descriptions}  ``translation" model $\translation(\desc | \rho, \library) \approx \probability[\desc | \program, \library]$ that describes how descriptions are generated for programs written in the library.

The concrete implementation described in the main paper uses a translation model that additionally  decomposes compositionally over language and programs--in particular, on the basis of  token-token translation distributions $\probability_\translation[w | l]$ between words $w \in \desc$ and $l \in \library$. Many available translation and semantic parsing models (such as synchronous grammars over natural language and programs) preserve this further compositional requirement (e.g. \cite{artzi2014learning,wong2006learning}). 

See Figure S\ref{joint-generative} (supplement) for example samples from the generative model on the graphics domain at earlier and later stages of training.

Our implementation uses a classical statistical machine translation model (the Model 4 version of the IBM Statistical Machine Translation models \cite{gal2013systematic}) whose parameters can be tractably estimated from very few paired programs and descriptions (in the distant supervision setting used in the original work, there may be no more than a couple of hundred training tasks in the full dataset, and fewer than 10 solved tasks on which to train the translation model at any given time.)  In addition to inference in small data settings, this translation model has a fully compositional generative definition \cite{gal2013systematic} that allows it to be easily used to train the neural amortized inference model which conditions on language. 

Despite this, however, this translation model (and the further inductive biases used to specifically relate program trees to sentences) make strong compositonality assumptions about the relationship between program primitives and words as a joint generative model of programs and language; we find that these inductive biases are useful in the small data setting and produce empirically successful results. However, this is likely because of \textit{how} the joint model is used during training, which does not require a perfect generative model of language (or language with respect to programs) for either amortizing inference or abstraction in order to use language as a heuristic during learning.

A full definition of the statistical translation model we use can be found in \cite{gal2013systematic}. We re-summarize important details here. The IBM family of translation models estimates the conditional token-token probabilities $\probability_\translation[w | l]$ on the basis of \textit{alignment} variables $a_{l,d}$, which specify a direct correspondence between tokens in parallel texts (e.g. a word in a task description and a program primitive.) These alignments are \textit{many:many} between tokens in programs and natural language sentences -- a given word can correspond to multiple primitives, and vice versa. Conditioned on a set of \textit{alignments} from paired programs and descriptions, the conditional probabilities in \textit{both} directions (the probability of generating a program primitive in a program based on the presence of a word in a sentence, and vice versa) are defined by marginalizing over the alignment variables. We provide one direction ($\probability_\translation[w | l]$), as the other is symmetrical:
$$
\probability_\translation[w | l] \propto \sum_{a_1}...\sum_{a_m} \probability[w, a_1...a_m | l] 
\propto \prod_{i=1}^{m}q(a_i | i, l, m)
$$
where $a_i$ are alignment variables inferred over a paired corpus and $q(j | i, l, m)$ can be interpreted as the probability of alignment variable $a_i$ (for the token with index $i$ in a program) taking value $j$ (where $j$ is an index into the corresponding sentence) conditioned on the lengths $l$ and $m$ of the program and natural language sentence \cite{gal2013systematic}. 

These alignments are inferred by approximately inverting the generative model in \cite{gal2013systematic} to maximize the likelihood of the observed paired sentences and programs. One implementation detail: the alignment algorithm operates over pairs of strings. For convenience we infer alignments between sentences and linearized token sequences in the program tree (which can be done with complete recoverability of the original program tree \cite{andreas2013semantic}). This is another inductive assumption that we choose after preliminary experimentation and find that our implementation yields strong empirical results regardless. 

The IBM translation model is a noisy-channel generative model that requires an additional language model $p(d)$ to generate language \cite{gal2013systematic,heafield2011kenlm}. We use an efficient parallelized implementation for inferring the translation model parameters from \cite{koehn2007moses}, which also contains a basic language model inference algorithm inferred over the full corpus of training task sentences (as a trigram model, which we again find simple but effective for our very small data setting). Specific model hyperparameters for all experiments are available in the released code repo (in the experiment runtime commands.)\\

\textbf{Mutual exclusivity}: Section 5.1 of the main paper also describes how the joint model can be modified to include language-specific priors, such as a simple implementation of the well-known \textbf{mutual exclusivity} prior documented in the cognitive language-learning literature \cite{markman1988children,gandhi2019mutual} and given a Bayesian formulation in \cite{frank2009using}. We provide an implementation to demonstrate that the joint model can be easily extended: specifically, a simple mutual exclusivity assumption can be added into the joint model by simply updating the compositional translation model to include additional distributions $t_{ME}(d_{new}| l)$ where $d_{new}$ are words that \textit{only} appear in unsolved training tasks and 
$$t_{ME}(d_{new}| l) \propto \alpha \probability[l | \library, \libparams]^{-1}$$
new words are now assumed to correspond to primitives \textit{inversely} proportional to their current usage under the learned program prior. As we show in the next section, incorporating this prior at the level of the joint model can be used to approximate mutual exclusivity assumptions in the learned search heuristic, encouraging exploration in the presence of new words.

Practically, we calculate the mutual exclusivity prior in our concrete implementation by leveraging the \textit{alignments} upon which our token-token translation probabilities are defined. Specifically, we add \textit{pseudoalignments} between each $d_{new}$ and each $l$  $\propto \alpha \probability[l | \library, \libparams]^{-1}$; when the token-token translation probabilities marginalize over the latent alignments and these pseudo alignments, the resulting translation probabilities encode the mutual exclusivity prior.

\subsection*{S5.2 Integrating the joint model into amortized conditional search}
\begin{figure} [h]
    \centering
    \includegraphics[width =0.45\textwidth]{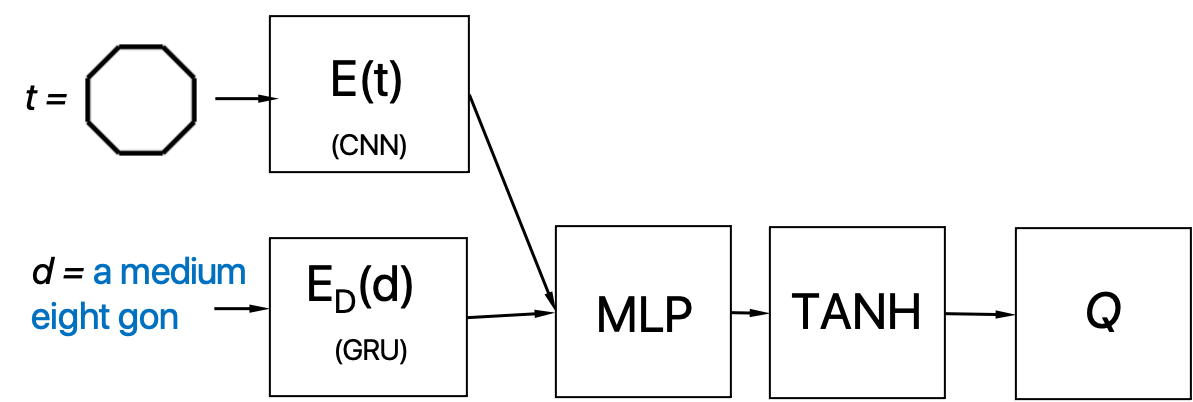}

    \caption{Architecture of the language-conditioned neural model $Q(\rho | d,t)$. The model takes as input task examples $t$. These are encoded using a domain-specific encoder $E(t)$. The model additionally takes in task descriptions $d$, encoded using a languag encoder $E_D(t)$ (implemented as a GRU).
    Task encodings are concatendated and feed to an MLP and activation layer and output a tensor $Q$. This parameterizes a distribution over program bigrams in the final DSL, which defines a conditional distribution from which to enumerate programs during search.}
     \label{laps-neural-model}
      
\end{figure}

The amortized conditional inference model $Q(\rho | t)$ (Sec. 4.2) extends straightforwardly in LAPS to condition on language  $Q(\rho | d,t)$ (Sec. 5.2). Importantly, the training procedure in Sec. 4.2 (training the neural model on samples from the prior) also extends to the language-enriched condition (training the neural model on samples from the joint prior, which include generated language annotations.)

In our experiments we implement the concrete neural model $Q(\rho | d,t)$ in our experiments by extending modularly on the original model in \cite{ellis2020dreamcoder} (and in the supplemental S4.2) for direct comparison.
Our full architecture therefore has \textit{three} modular components to additionally condition on language:
\begin{enumerate}
    \item A \textit{natural language task descriptions encoder} $E_D(d)$. This receives the task description $d$ as input.  We implement this as an RNN model using a bidirectional GRU \cite{cho2014learning} with 64 hidden units; we embed natural language symbols as 64-dimensional vectors, and randomly initialize and backpropagate through the embedding during training. We tokenize the sentences in $u$ on whitespace and concatenate each sentence, delimited by special start and end of sentence tokens. At test time, we replace any OOV tokens with a special UNK token.
    \item  A domain-specific task encoder $E(t)$, following S4.2.
    \item A bigram transition model over program primitives, following S4.2. To condition jointly on $E_D(d)$ and $E(t)$ we simply concatenate these two embeddings and update the first layer of the MLP to take the 128-dimensional concatenated embeddings as input.
\end{enumerate}

\subsection*{5.3 Abstraction learning as joint model compression}
Finally, the \textit{abstraction learning} model in \cite{ellis2020dreamcoder} can also be generalized to condition on language, by extending the optimal library inference algorithm with respect to the program prior to an optimal library inference algorithm with respect to the joint model over language and programs (Eq. 6 and 7, main text.)

In our concrete implementation with respect to the DreamCoder algorithm, this means extending the description-length compression objective -- originally defined over the program library and training task programs -- to include the translation model definition. The main paper defines a description-length prior over the compositional translation model (Eq. 10). Optimizing this tractably requires redefining the abstraction algorithm in \cite{ellis2020dreamcoder} -- which refactors $\lambda$-calculus programs via $lambda$-abstraction (see S4.3 for a summary) -- to also jointly re-estimate the description length of the translation model $\translation(\desc | \rho, \library')$ using the refactored programs under the new candidate library $\library'$. 

We implement an efficient approximation that can be calculated with respect to the classical statistical translation model described in S4.1 \cite{gal2013systematic}. In particular, we leverage the \textit{alignment}-based definition (which uses latent correspondences inferred between program tokens and sentence tokens in paired programs and descriptions) to approximate $-H(\probability_\translation[w|l]) = -\log(\probability_\translation[w|l]) $, the entropy of the token-token translation probabilities.

Specifically, as the IBM model defines the conditional token-token probabilities
$$
\probability_\translation[w | l] \propto \sum_{a_1}...\sum_{a_m} \probability[w, a_1...a_m | l]
$$
marginalized over alignments, where (slightly abusing notation) in any given paired program and sentence description we will have estimated a set of alignments $a_{w_j,l_k...l_n}$ between the $j$-th token in the description corresponding to one \textit{or more} tokens $l_k...l_n$ in the paired program.
 We therefore define the \textit{description}-length of each token-token translation as the sum of the description lengths of the alignments which express it under a library $\library$:

$$
\sum_{a_i}...\sum_{a_m} \probability[d, a_1...a_m | l, \library]  \propto \sum_{a_1}...\sum_{a_m} |a_i|_{\library}
$$
and the description lengths under the \textit{refactored} library $\library'$ containing new abstractions compresses according to 
\begin{equation}
\begin{split}|a'_{w_j,l'_k...l'_n}|_{\library'} <  |a'_{w_j,l_k...l_n}|_{\library} \iff
\\
\{l'_i \text{contains only } l_k...l_n \text{ as subcomponents}  | l'_k...l'_n\}
\end{split}
\end{equation}
and we say that a primitive $l \in \library$ is a \textit{subcomponent} of a refactored abstraction $l \in \library$ if the abstraction can be $\beta$-reduced such that $l$ appears in it. That is, a refactored alignment $a’ : w_i \to \{l’...l_n\}$ is compressed only when a new abstraction $l'$ encapsulates over a strict subset of the constituent program primitives already aligned to the word in the original alignment.
This allows us to re-approximate the description length of the new translation model with respect to a semantically-equivalent program refactoring without inducing $\probability_\translation[w | l]$ from scratch (which would require retraining the full translation model over the sentences and refactored programs.)

\section*{S6. Experiments}

This section describes additional details on each of the domains -- \textit{string editing, compositional graphics,} and \textit{scene understanding} -- in Section 6 of the main paper (see \textbf{Figure 2, main text} for examples from all three domains, shown along with the synthetic and human language annotations). We also provide additional details on the model and baseline hyperparameters available for each domain. All datasets generated for these experiments (including human language annotations) are released and links to static repositories are provided in the code release. We also release a complete set of commands to exactly replicate all model experiments.

All experiments for were conducted on a high-powered computing cluster using a fixed training budget of wall-clock search time per task for all models and baselines in a given domain (determined via hyperparameter search using the baseline model per domain, and reported on a per-domain basis below). The experiments on the string editing and graphics domains used models trained using 48 CPUs for search (using the original parallel enumerative search implemented in the released code for the DreamCoder model in \cite{ellis2020dreamcoder}); and the experiments trained on the scene reasoning task used 24 CPUs (as preliminary experiments revealed that these experiments required shorter search time for our main model, and we wished to reduce the carbon footprint of the remaining experiments after our first two domains.)

For all experiments we train the neural models for 1 $\times 10^4$ gradient steps. For experiments with language-guided compression, we use an upper bound of 5 new abstractions introduced per iteration. For mutual exclusivity experiments, we set $\alpha_{ME} = 0.1$. For all experiments, during program-only compression (see \cite{ellis2020dreamcoder} for a discussion of program-only compression hyperparameters) we use the hyperparameters from \cite{ellis2020dreamcoder} for parsimony with earlier work: a structure penalty of 1.5 and pseudocounts = 30.

\subsection*{S6.1 Domains}
(See \textbf{Figure 2}, main text for examples from all three domains, shown along with the synthetic and human language annotations.)
As discussed in the main paper, each domain consists of a dataset of \textit{tasks}; a set of procedurally generated \textit{synthetic language annotations}; and a set of \textit{human language annotations} provided by Mechanical Turk workers; we also described the \textit{base primitives} $\library_0$ with which all models (including baselines and ablations) were initialized for each domain.

\subsubsection*{S6.1.1 String Editing}
\textbf{Tasks:} structured string transformation problems taken from a publicly released dataset in \cite{andreas2017learning} (n=1000 train; n=500 test). Tasks consist of input dictionary strings transformed using randomly sampled regular expression transducer (n=30 examples per task). Transducers were sampled according to  abstract templates defined in \cite{andreas2017learning} and required identifying matched sequences of characters and \textit{adding} letters before them; \textit{removing} sequences; \textit{replacing} them with new sequences, or \textit{doubling} the sequence each time they appeared (See \textbf{Figure 2A, main text}).

\textbf{Language data:} The human language dataset for this domain was previously collected by \cite{andreas2017learning}. We defined a synthetic grammar of high-level templates over the ground truth regular expression transducers (corresponding to the original templates used to generate the tasks.) The synthetic templates were defined based on language from the original human annotations, and in most cases closely matched the true human provided annotations (which were generally quite structured), though with significantly less variation (the original language contained multiple human descriptions per task. We generate a single synthetic for each one. The synthetic dataset has a vocabulary size of n=44 for both train and test. We use the human annotations in the original dataset when evaluating on human data, which have a vocabulary of n=727 (train) and n=622 (test).)
We generate a synthetic dataset on this domain partly because of inaccuracies noted in \cite{andreas2017learning}.  The released code contains the complete generation procedure for these synthetic annotations. See Figure 2A for representative tasks with examples, synthetic language, and human descriptions.

\textbf{Initial program primitives:} 
We initialize all models with a set $\library_0$ of LISP-like primitives that operate over substring sequences to both construct regular expression match sequences and manipulate strings, augmented with three text manipulation-specific primitives intended for executing constructed regular expression sequences; \texttt{t} is a polymorphic type variable using standard Hindley-Milner polymorphism typing \cite{pierce}. The execution engine does include a regex-matching model; however, the synthesis model is naive to this execution engine and simply searches for manipulations over the input strings and the regexes as data arrays.

$\library_0$ contains 14 substring manipulation primitives, given below with type information. We also give a semantic gloss for primitives that are not standard LISP primitives.

\begin{itemize}
    \item \texttt{if (bool $\to$ t $\to$ t $\to$ t)}
    \item \texttt{cons (t $\to$ list(t) $\to$ list(t))} 
    \item \texttt{car (list(t) $\to$ t)}
    \item \texttt{cdr list(t) $\to$ list(t}
    \item \texttt{map ($(t_0 \to t_1) \to list(t_0) \to list(t_1)$)}
    \item  \texttt{tail (list(t) $\to$ t)} 
    \item \texttt{append (t $\to$ list(t) $\to$ list(t))} \\
    Appends element to end of list.
    \item \texttt{revcdr (list(t) $\to$ list(t))} \\
    Takes all except the last element of the list.
    \item \texttt{match (substr $\to$ substr $\to$ bool)} \\
    Returns true if the first argument, when executed as a regular expression, matches the second argument. 
    \item \texttt{regexsplit (substr $\to$ fullstr $\to$ list(substr))} \\
    Attempts to execute the first argument as a regular expression, and splits the second argument into a list of substrings, using the regular expression match as a delimiter (and includes the matched sequences in the returned list.) 
    \item \texttt{flatten (list(substr) $\to$ fullstr)}\\
    Flattens a list of substrings back into a string.
    \item \texttt{rconcat (substr $\to$ substr $\to$ substr)}\\
    Concatenates two substrings.
    \item \texttt{rnot (substr $\to$ substr)}\\
    Takes a substring argument \texttt{s} and returns the substring literal [\^\ s]
    \item \texttt{ror (substr $\to$ substr $\to$ substr)}
    \\Takes substring literals a and b and returns the substring literal ((a)|(b))
\end{itemize}
We also include 26 character constants of type \texttt{substr} and constants \texttt{dot} (regular expression wildcard character) and \texttt{empty} (empty string).

\textbf{Domain hyperparameters} We largely follow prior work \cite{ellis2020dreamcoder} to set algorithm training parameters; the earlier \cite{ellis2020dreamcoder} uses a 720s enumerative search budget for solving both text editing and general list manipulation tasks. We use the same 720s enumerative budget here.

The encoder E(t) follows the domain-specific encoder used for text and list editing problems in \cite{ellis2020dreamcoder}, a 2-layer GRU with 64 hidden units. The model is trained for a fixed gradient step budget (10,000 gradient steps) and we sample equally at random between supervision on the solved training tasks (and their solution programs in the current DSL) and samples from the joint generative model. As with \cite{ellis2020dreamcoder}, when generating tasks from the generative model, we use randomly sample inputs (on which we execute generated programs to produce an output.) 

\subsubsection*{S6.1.2 Compositional Graphics}
\textbf{Tasks:} inverse graphics problems (n=200  train; n=111 test) where each synthesis problem is specified by an image and solved by synthesizing a program in LOGO Turtle graphics \cite{abelson1986turtle}. The domain is inspired by the graphics domain in \cite{ellis2020dreamcoder} but intentionally re-designed to be much more challenging (ground-truth programs are much longer on average in the base programming language) and explicitly compositional: the training and testing tasks contain \textit{simple shape tasks} defined by compositional parameters for a set of basic shapes (\textit{a small triangle, a medium square; a small semicircle}); \textit{complex shape tasks} that require inferring more challenging (and longer) parameterized shapes (\textit{a greek spiral with eight turns}); and \textit{compositional tasks} defined by geometric rules and relations over the simple shapes (\textit{a seven sided snowflake with a short line and a small triangle as arms; a small triangle connected by a big space from a small circle}) (See \textbf{Figure 2C}).

\textbf{\textit{Simple parameterized shapes}} are either polygons (\textit{triangle, square, [n] gon}), curves (\textit{semicircle, circle}) or \textit{line}s. Simple shapes are parameterized by one of three sizes (\textit{small} or \textit{short}; \textit{medium}; and \textit{big}). When generating synthetic language descriptions, pluralized objects are tokenized with separate tokens for the noun lemma and a token for the plural suffix (e.g. \textit{square s}). \\
\textbf{\textit{Complex parameterized shapes}} require constructing more complex images out of basic lines, and are intended to evaluate performance on tasks that pose a greater search challenge in the initial DSL, and whose structure is not directly cued by compositional relationships over easier components. Further, the complex shapes can be solved using abstractions (e.g. for repeatedly rotating a pen at right angles) that are not directly cued by shared lexical names -- we evaluate the algorithm's ability to learn and use abstractions that correspond to useful sublexical structures shared across multiple lexemes. We define four template families for complex shapes: \textit{spiral}s, \textit{staircase}s, \textit{zigzag}s, and \textit{star}s. \\
\textbf{\textit{Compositional graphics}} tasks invoke compositional relationships over the simple parameterized shapes. We define templates for generating 6 families of compositional tasks: \textit{nested}, \textit{next to}, \textit{separated by}, \textit{connected by}, \textit{in a row}, and \textit{snowflake}s. 

\textbf{Language data:}  
We gather human language annotations by asking Mechanical Turk workers to write an image description for the rendered graphics images that specify each task. Each worker labeled 20 training and 10 testing images after viewing a disjoint, randomly sampled set of 15 example images paired with their synthetic language captions. (Workers were asked to write a \textit{short, clear description that a person or robot could use to recreate the picture}, and told that the examples were paired with \textit{automatically generated captions as an example of the kinds of descriptions you could write for this picture}.) We control for description quality by requiring workers to complete a reference task on their own descriptions: after writing their initial annotations, workers were required to correctly match each annotation to the target image (from amidst a set of 12 distractors drawn heuristically from similar images on the full task dataset, and other images they themselves had described), and only annotations correctly matched to the target image were retained (workers were given a chance to redescribe pictures they failed to match to their own captions.) We preprocess the human dataset minimally to standardize number terms (e.g. we use the same token type for both \textit{3} and \textit{three}) and to split plurals into a lemma and suffix, as in the synthetic dataset. The final dataset has a vocabulary size of n=562 for both train and test.

As with the string editing domain, we define a synthetic dataset using parameterized templates based on systematic language reused in the human annotations (see Figure 2A for a comparison between human annotations and synthetic language); as with that domain, we choose a synthetic dataset to ensure systematic re-use of high level terms for repeated compositional objects (such as the ``n-gon" or ``snowflake" terminology.)

We then generate graphics tasks by defining parameterized templates over ground truth programs \textit{in $\library_0$}, and a corresponding generator for synthesizing natural language descriptions based on each ground truth program. It is important to note that the templates are defined at any extremely high level and were written with respect to low-level programs in a simple graphics language (many of which were derived by generalizing compositionally over complex structures in \cite{ellis2020dreamcoder}, such as the `snowflake' images).

\textbf{Initial program primitives:} 
For comparison with prior work, our initial library on this domain (and the base language used to generate the ground truth graphics programs) is an implementation of the LOGO Graphics DSL used in \cite{ellis2020dreamcoder}, which consists of four typed, imperative primitives modeled within the $\lambda-$calculus with a state monad $S$: 
  
\begin{tabbing}
  \texttt{move: distance $\to$ angle  $\to$ S  $\to$ S} \\ 
  \texttt{pen-up: (S  $\to$ S)  $\to$ S  $\to$ S} \\ 
  \texttt{for: int  $\to$ (S  $\to$ S)  $\to$ S  $\to$ S} \\ 
  \texttt{get/set: (S  $\to$ S)  $\to$ S  $\to$ S} \\
\end{tabbing}
as well as four arithmetic operators (+, -, *. /), integer constants (1-9), unit distances and angles (1 meter and $2\pi$ radians), and special values $\infty$ and $\epsilon$.

Figure 3 (main text) shows examples of the graphics tasks, synthetic descriptions, human descriptions, and sample programs in the ground truth initial DSL.

\textbf{Domain hyperparameters}
We largely follow prior work \cite{ellis2020dreamcoder} to set algorithm training parameters. Consistent with the graphics program experiments in \cite{ellis2020dreamcoder}, we train all models, including baselines and ablations, using an enumerative search budget of 1800s per task (both when using pure enumerative search from the DSL prior, and neurally-guided search conditioned on the task examples and language descriptions); the results in Table 1 compare the relative advantage of our model given this fixed search time. We train all models on 48 CPUs during parallel enumerative search, and run the algorithm for a maximum of 27 iterations (see learning curves. As we run multiple random seed replications of models in this domain, we tuned the iteration limit based on performance on the first replication, allowing models models to train while performance continued to increase. To conserve computational resources, we later stopped several of our own model replications before 27 iterations, as they had reached near ceiling performance. As we report the best held-out test score across all 27 iterations for any one model, the early stopping would only serve to give a conservative estimate on performance for these models.) We randomly reorder the training set of tasks once before the first loop, then iterate through batches of n=40 tasks at each iteration; learning curves show results from evaluating on held-out tasks every n=3 iterations. 

The encoder E(t) follows the domain-specific encoder used for the original graphics domain in \cite{ellis2020dreamcoder} for a more direct comparison: we use a 6-layer CNN, where each layer consists of a 64x64 2D convolutional sublayer with kernel size = 3, a RELU activation sublayer, and a max-pooling sublayer with kernel size = 2.  The model is trained for a fixed gradient step budget (10,000 gradient steps) and we sample equally at random between supervision on the solved training tasks (and their solution programs in the current DSL) and samples from the joint generative model.

\subsubsection*{S6.1.3 Scene Reasoning}
\textbf{Tasks:} inductive scene reasoning tasks (n= 212 train; n=115 test) where each synthesis problem is specified by a structured input scene, and outputs can be a number (\textit{how many red rubber things are there?}), a boolean value (\textit{are there more blue things than green things?}), or another scene (\textit{what if all of the red things turned blue?}). This domain is modeled on CLEVR \cite{johnson2017inferring} but designed to support non-linguistic, inductive synthesis in the programming-by-example paradigm: each task is specified with \textit{n=7} paired input output examples. See \textbf{Figure 2B, main text} for example tasks showcasing the original and extended templates, synthetic language annotations, and human language annotations.

The dataset includes questions randomly generated from the following subset of the \textit{original CLEVR question templates} (see \cite{johnson2017inferring} for additional details on the task generation process and question templates; we also release our own augmented question generation code and the full dataset):
\begin{itemize}
    \item \textbf{zero\_hop}: questions that require counting or answering an attribute query about a subset of objects in the scene. (e.g. \textit{How many small cylinders are there?}; \textit{What material is the purple thing?}).
    \item \textbf{one\_hop}: questions similar to the \textit{zero\_hop} tasks, but that require reasoning over an additional relational query (e.g \textit{What number of things are right the small gray thing?}).
    \item \textbf{single\_or}: questions that additionally introduce a \textit{disjunction} between sets of objects. (e.g. \textit{How many objects are either large metal spheres or large rubber things?})).
    \item \textbf{(compare\_integer}: questions that additionally introduce a $\geq$ or $\leq$ operator between counts of sets of objects. (e.g. \textit{Is the number of large rubber cubes less than the number of large green rubber things?})
    \item \textbf{same\_relate}: questions that additionally require reasoning about other objects with the same attribute as a specified object. (e.g. \textit{How many other things are there of the same size as the cyan thing?}).
\end{itemize}
We choose these templates as a representative subset of the style of the full CLEVR dataset, that requires the full language of high-level primitives in \cite{johnson2017inferring} to solve. We omit some longer questions in the same format (e.g. \textit{two\_hop}) as our intention is to compare synthesis baselines, rather than to achieve SOTA performance on CLEVR: this would likely only increase the computing resources needed to compare the various methods and we already found a significant differential between our model and the baselines on the shorter questions.)

We also add \textit{new} question templates generated in the style of the original CLEVR tasks, but designed to model other common AI tasks (such as generating new scenes based on existing ones) and to require new abstractions (that were not expressible in the original restricted symbolic language used to generate scenes in \cite{johnson2017inferring}):
\begin{itemize}
    \item \textbf{localization}: questions for object localization. These return an output \textit{scene} consisting of a localized set of objects based on a set of query attributes (e.g. \textit{Find the gray rubber thing.}).
    \item \textbf{remove}: questions that either return an output \textit{scene} with a subset of the objects removed, or that query about latent scenes where a subset of objects has bee removed. (e.g \textit{What if you removed all of the gray metal things?}; \textit{If you removed the green cubes, how many cubes would be left?}).
    \item \textbf{transform}: questions that either return an output \textit{scene} where a subset of the objects has been \textit{transformed} to set new attributes, or that query about latent scenes where a subset of objects has been modified this way. (e.g \textit{What if all the blue metal things became rubber things?}; \textit{If all of the large yellow rubber things became gray spheres, how many gray spheres would there be?}).

\end{itemize}
We treat these as program synthesis tasks: the input scenes are specified as \textit{symbolic scene graphs consisting of an array of structured, objects defined as a dictionary of their attributes}, and programs are designed to manipulate these structured arrays (this data structure is the original format in which scenes themselves are generated in \cite{johnson2017inferring}; the images displayed in Figure 3, main text are rendered using the original image rendering pipeline). Our intention is \textit{not} to build a visual reasoning architecture: rather, we are interested in learning structured manipulations of scenes. We see work in \textit{inverse graphics} (such as \cite{yi2018neural}) which outputs a structured scene graph based on pixel images as the \textit{first} step in a symbolic processing and reasoning pipeline as analogous; we are interested in the structured manipulation of these scene representations.

\textbf{Language data:} Synthetic language annotations are generated based on the original high-level templates in \cite{johnson2017inferring}, as well as additional templates we define for the extended questions in the same style. We gather human language annotations by asking Mechanical Turk workers to write an instruction or question describing the set of inductive examples. However, due to the difficulty of solving certain tasks in a limited time frame based on the inductive examples alone (such as the questions about disjunctions over scenes), we show Mechanical Turk workers the synthetic descriptions for this domain and ask them to write a semantically similar description that changes more than one word in the original caption, and that would be "more natural for a human to understand". This paraphrasing paradigm is similar to that used in \cite{wang2015building}, though we find that in comparison to other domains it generates less diverse language data.) We remove all punctuation, tokenize on spaces, and use an additional domain heuristic to stem all plurals (e.g. \textit{cubes}).

\textbf{Initial program primitives:} We initialize all models with a set $\library_0$ of LISP-like primitives. These are similar to the initial list manipulation primitives used in the \textit{string editing} domain: as both domains can be treated as manipulating structured arrays, we are interested in learning differentiated, domain-specific abstractions based on a very similar base language. $\library_0$ also includes primitives for querying attributes of objects on the domain (these are typed getters that simply query the object dictionary of attributes) and several domain-specific functions necessary for manipulating these attribute. We deliberately use a much more base level programming language than the high-level, domain-specific language hand-designed in \cite{johnson2017inferring}; our goal is to \textit{learn} the necessary abstractions.

We give a semantic gloss for primitives that are not standard LISP primitives.
\begin{itemize}
    \item \texttt{if (bool $\to$ t $\to$ t $\to$ t)}
    \item \texttt{cons (object $\to$ list(object) $\to$ list(object))} 
    \item \texttt{car (list(object) $\to$ object)}
    \item \texttt{map ($(t_0 \to t_1) \to list(t_0) \to list(t_1)$)}
    \item \texttt{fold ($(list(t) \to list(t)) \to (t \to list(t) \to list(t)) \to list(t)$)}
    \item \texttt{len (list(t) $\to$ int)}
    \item \texttt{$>$ (list(t) $\to$ bool)}
    \item \texttt{$<$ (list(t) $\to$ bool)}
    \item \texttt{set\_union (list(t) $\to$ list(t) $\to$ list(t))}
    \item \texttt{set\_intersect (list(t) $\to$ list(t) $\to$ list(t))}
    \item \texttt{set\_difference (list(t) $\to$ list(t) $\to$ list(t))}
    \item \texttt{relate (object $\to$ relation $\to$ list(t))} Returns an array of objects that satisfy a spatial relation with respect to an input object.

\end{itemize}
We also include \textit{equality} comparators for each of the attribute types (e.g. \texttt{eq\_color?}; \textit{getters} for each attribute, and \textit{setters} for each attribute. We also include integer constants 0-9 for counting and constants for the attributes (\texttt{blue, red, big, small, rubber, metal}) based on the original object and spatial relation constants \cite{johnson2017inferring}.\\
\textbf{Domain hyperparameters:} We run a coarse hyperparameter search based on the baseline model to set the domain hyperparameters. We train all models, including baselines and ablations, using an enumerative search budget of 1000s per task and run the models for a maximum of 5 iterations. we run multiple random seed replications reordering the training set, in the same way as the compositional graphics domain. The results in Table 1 also compare a \textit{curriculum} ordering of the training set based on the number of tokens in the synthetic language captions (split on spaces.)

The encoder E(t) is a variant of the RNN-based domain-specific encoder used for text and list editing problems in \cite{ellis2020dreamcoder} (as well as the string editing domain). The model is trained for a fixed gradient step budget (10,000 gradient steps) and we sample equally at random between supervision on the solved training tasks (and their solution programs in the current DSL) and samples from the joint generative model. As with \cite{ellis2020dreamcoder}, when generating tasks from the generative model, we use randomly sample inputs (on which we execute generated programs to produce an output.) We encode the symbolic scene data structures with the RNN by encoding a flattened version of the scene graph. The scene graph is originally stored as a dictionary of attributes; when flattened, we indicate the dictionary structure using special tokens to denote the keys and the start and end of any array delimiters (the original scene graph is fully reconstructable from the flattened version.)

\subsection*{S 6.2 Results and Additional Qualitative Results}
In this section, we discuss additional qualitative results from an in depth exploration of the graphics domain that were omitted from the main paper for space, but provide additional insight on the behavior of the learned model in the hardest learning domain (based on the differential between baseline and LAPS-augmented performance.)

\textbf{Learned abstractions and synthesized programs.} Figure S\ref{graphics-abstractions-extended} (supplement) show sample abstractions in the final libraries $\library_f$ for the best performing models in the graphics domain as a concrete exemplar of abstractions that are learned and how they are used, along with sample tasks solved with these abstractions. The figures are shown as dependency graphs to indicate how progressively more complex abstractions \textit{build} on abstractions at prior iterations of learning; we also show selected probabilities from the translation model (depicted are examples from the top-3 primitive translations for a given word; some primitives are not high probability translations for any word.)

\begin{figure*}
    \centering
    \includegraphics[width =\textwidth]{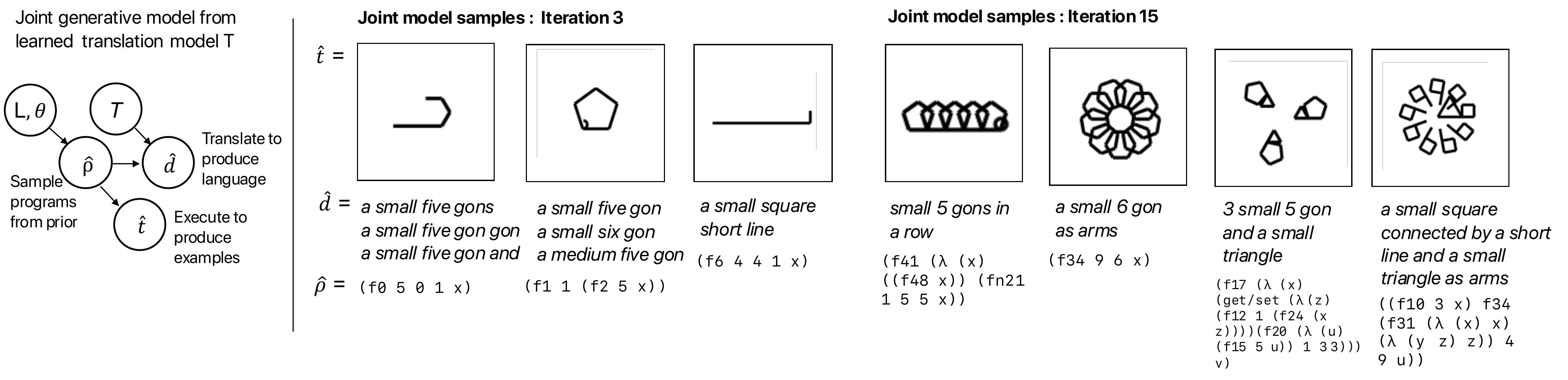}
    \vspace{-5mm}
    \caption{(left) Joint generative model J over programs sampled from the DSL prior and natural language produced by the translation model $T(D | \library)$, inferred from  
    solved training tasks. Samples from the model are used to train a neural synthesizer to guide search on more challenging, unsolved tasks. (right) Samples from the $J$
    generative model in the graphics  domain shows how program complexity increases and generated language improves across iterations, as the system both adds richer abstractions to the DSL and learns better alignments over the solution set, enabling the trained neural model to solve more complex tasks}
    \label{joint-generative}. 
\end{figure*}

\begin{figure*}
 
    \centering
    \includegraphics[width =\textwidth]{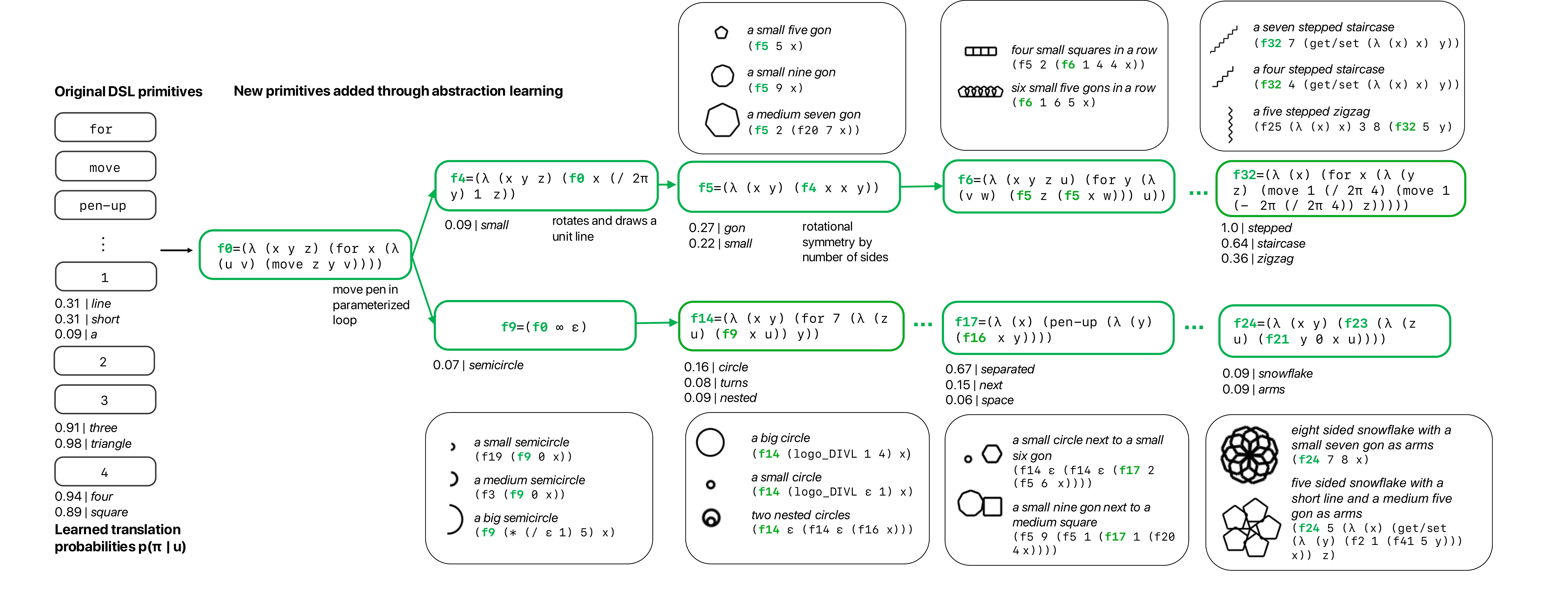}
    \vspace{-3mm}
    \caption{Abstractions and programs learned for the graphics domain. Sample abstractions (right) learned from a minimal starting DSL (left) for solving progressively more complex graphics program synthesis tasks with language annotations.  Also shown with translation probabilities. Our iterative algorithm learns alignment-based translation probabilities between natural language words and program primitives to guide program search and abstraction (depicted are examples from the top-3 primitive translations for a given word; some primitives are not high probability translations for any word.}
     \label{graphics-abstractions-extended}
      
\end{figure*}

\textbf{Joint generative model samples.} Figure S\ref{joint-generative} (supplement) shows samples from the joint generative model on the graphics domain (programs from the library which are executed to produce the task example image, and translated to produce language annotations) at early and later stages of training, indicating that the joint model itself improves as learning improves, which itself allows better training for the conditional inference model and better abstraction guiding based on language.

\bibliography{icml_2021_supplemental}
\bibliographystyle{icml2021}